\documentclass{bmvc2k}

\usepackage[utf8]{inputenc} 
\usepackage[T1]{fontenc}    
\usepackage{url}            
\usepackage{booktabs}       
\usepackage{amsfonts}       
\usepackage{nicefrac}       
\usepackage{microtype}      
\usepackage{xcolor}     
\usepackage{graphicx}
\usepackage{amsmath}
\usepackage{amssymb}
\usepackage{multirow,multicol}
\usepackage[linesnumbered,ruled,vlined]{algorithm2e}
\usepackage{siunitx}
\usepackage{bbm}
\usepackage{diagbox}
\usepackage{colortbl}
\usepackage{float}
\usepackage{wrapfig}
\usepackage{pifont}
\usepackage{csquotes}
\definecolor{ForestGreen}{RGB}{34,139,34}
\colorlet{LightGreen}{ForestGreen!60}

\title{\textsc{Retro}: \underline{Re}using \underline{t}eacher p\underline{ro}jection head for efficient embedding distillation on Lightweight Models via Self-supervised Learning}

\addauthor{Khanh-Binh~Nguyen}{http://www.ncc.re.kr}{1}
\addauthor{Chae~Jung~Park}{http://www.ncc.re.kr}{1,2}

\addinstitution{
 Cancer Big Data and AI\\
 National Cancer Center\\
 Goyang, South Korea
}
\addinstitution{
 Corresponding Author
}

\runninghead{Khanh-Binh~Nguyen, Chae~Jung~Park}{\textsc{Retro}}

\begin{document}

\maketitle

\begin{abstract}
    Self-supervised learning (SSL) is gaining attention for its ability to learn effective representations with large amounts of unlabeled data. 
    Lightweight models can be distilled from larger self-supervised pre-trained models using contrastive and consistency constraints, but the different sizes of the projection heads make it challenging for students to accurately mimic the teacher's embedding. 
    We propose \textsc{Retro}, which reuses the teacher's projection head for students, and our experimental results demonstrate significant improvements over the state-of-the-art on all lightweight models. 
    For instance, when training EfficientNet-B0 using ResNet-50/101/152 as teachers, our approach improves the linear result on ImageNet to $66.9\%$, $69.3\%$, and $69.8\%$, respectively, with significantly fewer parameters.
\end{abstract}

\section{Introduction}
Deep learning has achieved remarkable success in various visual tasks, such as image classification, object detection, and semantic segmentation, thanks to the availability of large-scale annotated datasets. 
However, acquiring labeled data is time-consuming and expensive, making it crucial to explore better ways to utilize unlabeled data. 
Self-supervised learning (SSL) has emerged as an effective method to learn useful representations on unlabeled data, resulting in an outstanding performance on downstream tasks \cite{gidaris2018unsupervised,noroozi2016unsupervised,doersch2015unsupervised,pathak2016context,chen2020simple,chen2020big,grill2020bootstrap,he2020momentum}.

Despite its effectiveness, most SSL methods require large networks, and the performance deteriorates when the model size is reduced. 
To address this issue, \citep{fang2021seed} proposed SEED, a self-supervised representation distillation method that distills the knowledge of larger pre-trained models into lightweight models via self-supervised learning. 
Similarly, CompRess \cite{abbasi2020compress} mimics the similarity score distribution between a teacher and a student over a dynamically maintained queue. 
\citep{gao2022disco} suggests incorporating consistency constraints between teacher and student embeddings to alleviate the Distilling Bottleneck problem via DisCo. 
BINGO \cite{xu2021bag} aims to transfer the relationship learned by the teacher to the student by leveraging a set of similar samples constructed by the teacher and grouped within a bag.

\begin{wrapfigure}{r}{0.5\textwidth}
  \centering
    \includegraphics[width=\linewidth]{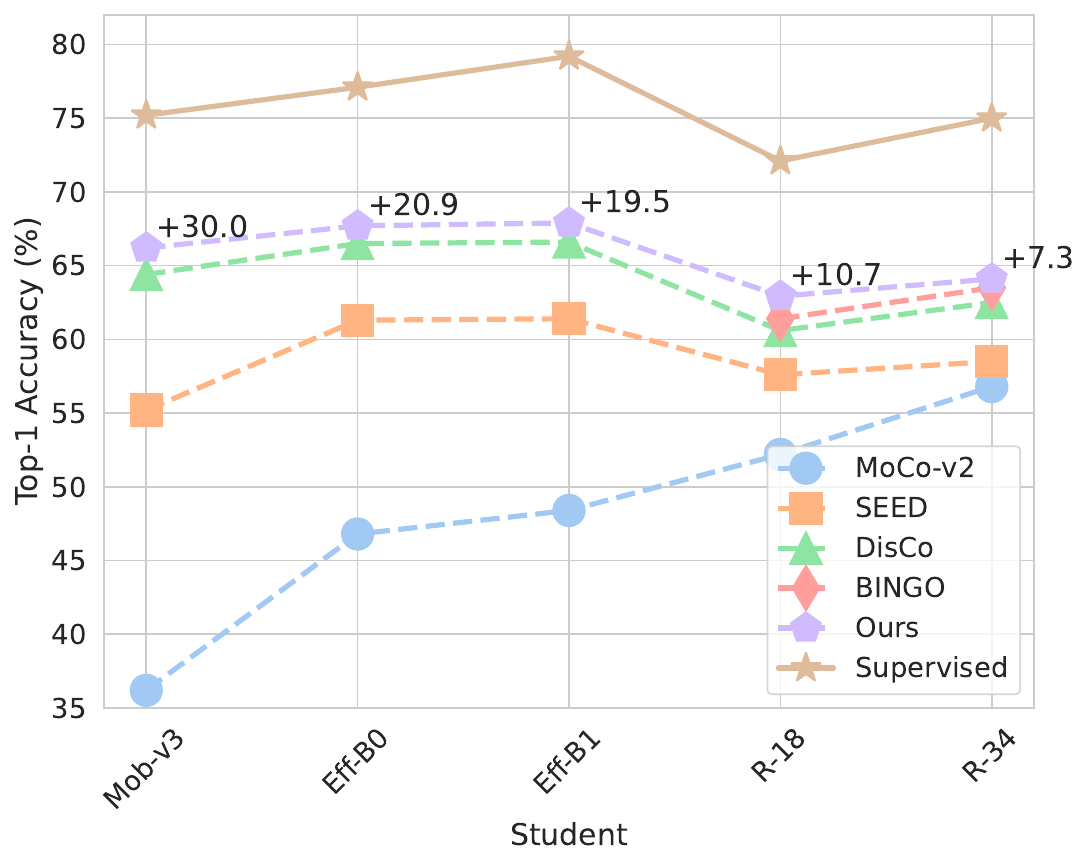}
    \caption{ImageNet top-1 linear evaluation accuracy on different network architectures.
    Our method significantly exceeds the result of using MoCo-V2 directly and surpasses the state-of-the-art DisCo by a large margin.
    Particularly, the result of EfficientNet-B0 is quite close to the teacher ResNet-50, while the number of parameters of EfficientNet-B0 is only 16.3\% of ResNet-50.
    The improvement brought by \textsc{Retro} is compared to the MoCo-V2 baseline.}
    \label{fig:quality-vs-quantity}
\end{wrapfigure}

Despite achieving state-of-the-art results across multiple tasks with high performance, some concerns remain. 
First, \cite{gao2022disco} discovered that expanding the dimension of the hidden layer in the MLP (projection head) could alleviate the Distilling Bottleneck problem. 
However, this approach is trivial since determining the size of the dimension and how large it should be remains unanswered. 
Second, because the student is lightweight with limited capability, it is challenging to accurately mimic the teacher from the encoder to the projection head. 
For example, in the DisCo study \cite{gao2022disco}, they expanded the dimension to $2048$, which is the projection head dimension of ResNet-50/101/152. 
Consequently, this approach is equivalent to increasing the capability of mimicking the teacher, resulting in improved performance. 
However, when using ResNet-50$\times$2 with a dimension of $8192$ as a teacher, the performance on MobileNet-v3-Large and EfficientNet-B1 drops significantly and is inferior to the previous method \cite{fang2021seed}. 
Moreover, the feature distributions of the teacher and student models are statistically different and cannot be directly compared in practice, even if their dimensions are the same. 
Therefore, the optimal dimension for the projection head and how to efficiently distill the teacher embedding remain unanswered questions.

In this study, we propose a novel approach for improving the performance of distilling lightweight models through SSL. 
Specifically, we suggest reusing the pre-trained teacher projection head for students, instead of mimicking it during training. 
This is based on the hypothesis that the most valuable knowledge is contained in the projection head, and it should be retained during distillation. 
Our proposed \enquote{teacher projection head reusing strategy} involves replacing the student projection head with the pre-trained one from a teacher, which is a large dimension MLP layer that has been optimized. 
This enables direct reuse of the projection head, without the need for heuristic selection of the dimension size via trial and error. 
Additionally, a \enquote{dimension adapter} is inserted between the student encoder and the teacher projection head to align the dimension.

Our approach simplifies the training objective from mimicking the representation and the embedding to aligning the representation with the optimal embedding. 
Our experiments show that the proposed method, named \textsc{Retro}, outperforms the existing DisCo method by a significant margin when using the same architecture on various downstream tasks. 
Moreover, \textsc{Retro} achieves state-of-the-art SSL results on all lightweight models, including ResNet-18/34, EfficientNet-B0/B1, and MobileNetV3. 
Notably, the linear evaluation results of EfficientNet-B0 on ImageNet are comparable to ResNet-50/ResNet-101, despite having only a fraction of the parameters. 
On the COCO and PASCAL VOC datasets, \textsc{Retro} also achieves more than $3\%$ mAP improvement across different pre-trained models.

\section{Related Work}
Self-supervised learning and knowledge distillation have emerged as crucial research areas in machine learning, attracting significant attention in recent years. 
In this section, we present a review of some of the key works in these fields.

\subsection{Self-supervised Learning}
Self-supervised learning is an essential branch of unsupervised learning that automatically generates supervisory signals from unlabeled data. 
One of the earliest and most effective techniques used in self-supervised learning is the autoencoder, which compresses the input data and reconstructs it. 
Contrastive learning is another popular self-supervised learning method that enables the model to differentiate between similar and dissimilar pairs of examples. 

Recent studies have demonstrated the efficacy of contrastive-based techniques in self-supervised representation learning, where different perspectives of the same input are encouraged to be closer in feature space \cite{chen2020simple,chen2020big,chen2021exploring,chen2020improved,grill2020bootstrap,he2020momentum}. 
For instance, SimCLR \cite{chen2020simple,chen2020big} has proved that using strong data augmentation, larger batch sizes of negative samples, and including a projection head (MLP) after global average pooling can boost self-supervised learning. 
However, the performance of SimCLR is dependent on very large batch sizes and may not be feasible in real-world scenarios. 

MoCo \cite{chen2020improved,he2020momentum}, on the other hand, uses a memory bank to maintain consistent representations of negative samples. 
It considers contrastive learning as a look-up dictionary, enabling it to achieve superior performance without large batch sizes, making it more practical. 
BYOL \cite{grill2020bootstrap} introduces a predictor to one branch of the network to prevent trivial solutions and break the symmetry. 
DINO \cite{caron2021emerging} applies contrastive learning to vision transformers with self-distillation intuition.

\begin{figure*}[!htb]
    \centering
    \subfigure[CompRess]{\label{fig:compress}\includegraphics[width=0.24\linewidth]{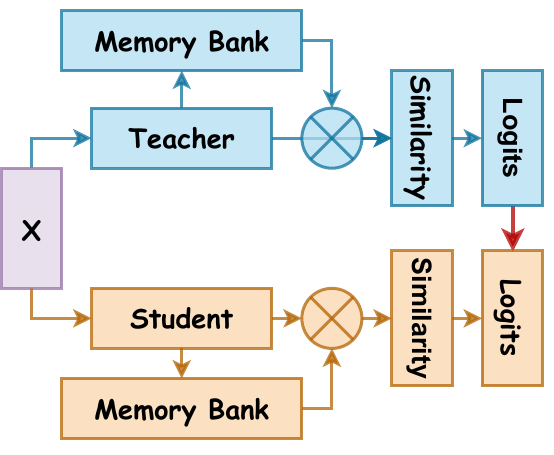}}
    \subfigure[SEED]{\label{fig:seed}\includegraphics[width=0.24\linewidth]{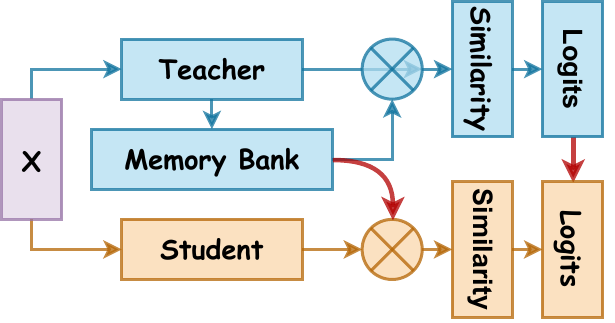}}
    \subfigure[DisCo]{\label{fig:disco}\includegraphics[width=0.24\linewidth]{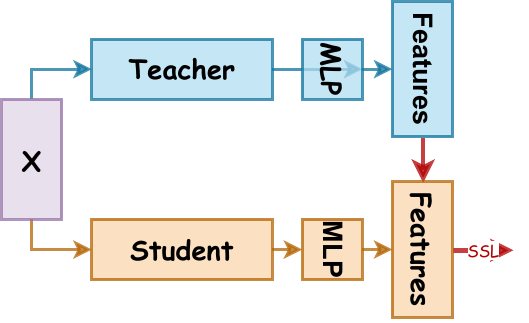}}
    \subfigure[Proposed]{\label{fig:proposed}\includegraphics[width=0.24\linewidth]{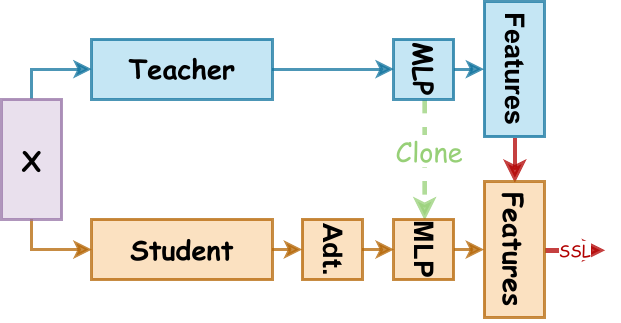}}
  
    \caption{Comparison with existing self-supervised distillers.
    $x$ is the input image.
    The orange arrow indicates the knowledge transfer direction.
    Both \ref{fig:compress} CompRess \cite{abbasi2020compress} and \ref{fig:seed} SEED \cite{fang2021seed} transfer the knowledge of the similarity between a sample and a negative memory bank.
    \ref{fig:disco} DisCo \cite{gao2022disco} constrains the last embedding of the student to be consistent with that of the teacher.
    \ref{fig:proposed} Our \textsc{Retro} improves DisCo by reusing the teacher projection head for the student, which has a higher capability to generate generalized embedding. 'Adt.' indicates the adapter layer.}
    \label{fig:comparision}
\end{figure*}

\subsection{Knowledge Distillation}
Knowledge distillation is a powerful technique used for transferring knowledge from a large, complex model (known as the teacher) to a smaller, simpler model (known as the student) to improve its performance. 

The idea of knowledge distillation was first proposed by \citep{hinton2015distilling}, which transfers knowledge from a large teacher to a smaller student by minimizing the Kulback-Leibler (KL) divergence between the outputs of the two models. 
Attention Transfer (AT) is introduced to transfer the spatial attention of the teacher to the student by minimizing the mean squared error (MSE) between the feature maps of the two models. 
This method guides the student to focus on relevant regions of the input image, improving its performance on small datasets.=

FitNets \cite{romero2014fitnets} is another method of knowledge distillation that transfers knowledge from the intermediate layers of a deep and thin teacher to a deeper but thinner student. 
The intermediate layers learned by the teacher are treated as hints, and the student is trained to mimic them using mean squared error loss.
Relation Knowledge Distillation (RKD) \cite{park2019relational} is a method that transfers the mutual relationship between the samples in a batch from the teacher to the student. 
RKD uses distance-wise and angle-wise distillation loss to transfer the relationship between the samples to the student.

\section{Method}
\subsection{Self-supervised Learning and Knowledge Distillation}
In recent years, there has been a growing interest in combining knowledge distillation and self-supervised learning to improve the learning process. 
Some recent works, such as CRD \cite{tian2019contrastive} and SSKD \cite{xu2020knowledge}, have used self-supervision as an auxiliary task to enhance knowledge distillation in fully supervised settings by transferring relationships between different modalities or mimicking transformed data and self-supervision tasks.

On the other hand, CompRess \cite{abbasi2020compress} and SEED \cite{fang2021seed} have focused on improving self-supervised visual representation learning on small models by incorporating knowledge distillation. 
They leverage the memory bank of MoCo \cite{he2020momentum} to maintain the consistency of the student's distribution with that of the teacher. 
Meanwhile, DisCo \cite{gao2022disco} proposes to align the final embedding of the lightweight student with that of the teacher, exploiting the student's learning ability to maximize knowledge. 
They also increase the dimension of the student's projection head to better mimic the teacher's embedding.

However, the questions of which knowledge is essential for the student and how to efficiently distill it remain unanswered. 
Moreover, previous approaches focused only on making the student mimic the teacher instead of exploiting the student's learning ability. 
Our proposed method aims to enhance the self-supervised representation learning ability of lightweight models by aligning the student encoder with the teacher's projection head instead of merely mimicking the teacher. 
Figure \ref{fig:comparision} illustrates the differences between our proposed method and CompRess, SEED, and DisCo.

In this section, we will provide a detailed description of our proposed method, \textsc{Retro}. 
We will start by reviewing the preliminary concepts of contrastive-learning-based SSL. 
Next, we will discuss the overall framework of \textsc{Retro} and explain how it works. 
Finally, we will introduce the objective of \textsc{Retro} and describe the process of updating its parameters.

\subsection{Preliminary on Contrastive Learning Based SSL}
\subsubsection{Contrastive Learning Based SSL}
In contrastive-learning-based SSL, the goal is to predict whether a pair of instances belong to the same class or different classes. 
The two instances are obtained by applying different data augmentation techniques to the same input image $x$, resulting in two augmented views $v$ and $v'$ of the same instance. 
The objective is to make the two views similar while views of different instances should be dissimilar. 
Each view is then passed through two encoders $f_s\left(\cdot\right)$ and $f_t\left(\cdot\right)$ to obtain the corresponding representations $z_s$ and $z_t$.

To map the high-dimensional representations to a lower-dimensional embedding, a projection head $g\left(\cdot\right)$, which is a non-linear MLP, is used. 
Specifically, $g$ takes the representation $z$ as input and maps it to an embedding $E$ as $E = g\left(z\right) = g \circ f\left(x\right)$. 
These embeddings are then used to estimate the similarity in contrastive learning. 
The projection head is crucial to the success of self-supervised learning, as demonstrated in prior works such as MoCo \cite{he2020momentum} and SimCLR \cite{chen2020simple}. 
The encoder can be any network architecture, such as ResNet or EfficientNet.

The projection head consists of two linear layers followed by a non-linear activation function such as ReLU.
The output dimension of the projection head is typically set to a smaller value, such as $128$, to obtain a low-dimensional embedding.

\begin{wrapfigure}{r}{0.55\textwidth}
  \centering
    \includegraphics[width=\linewidth]{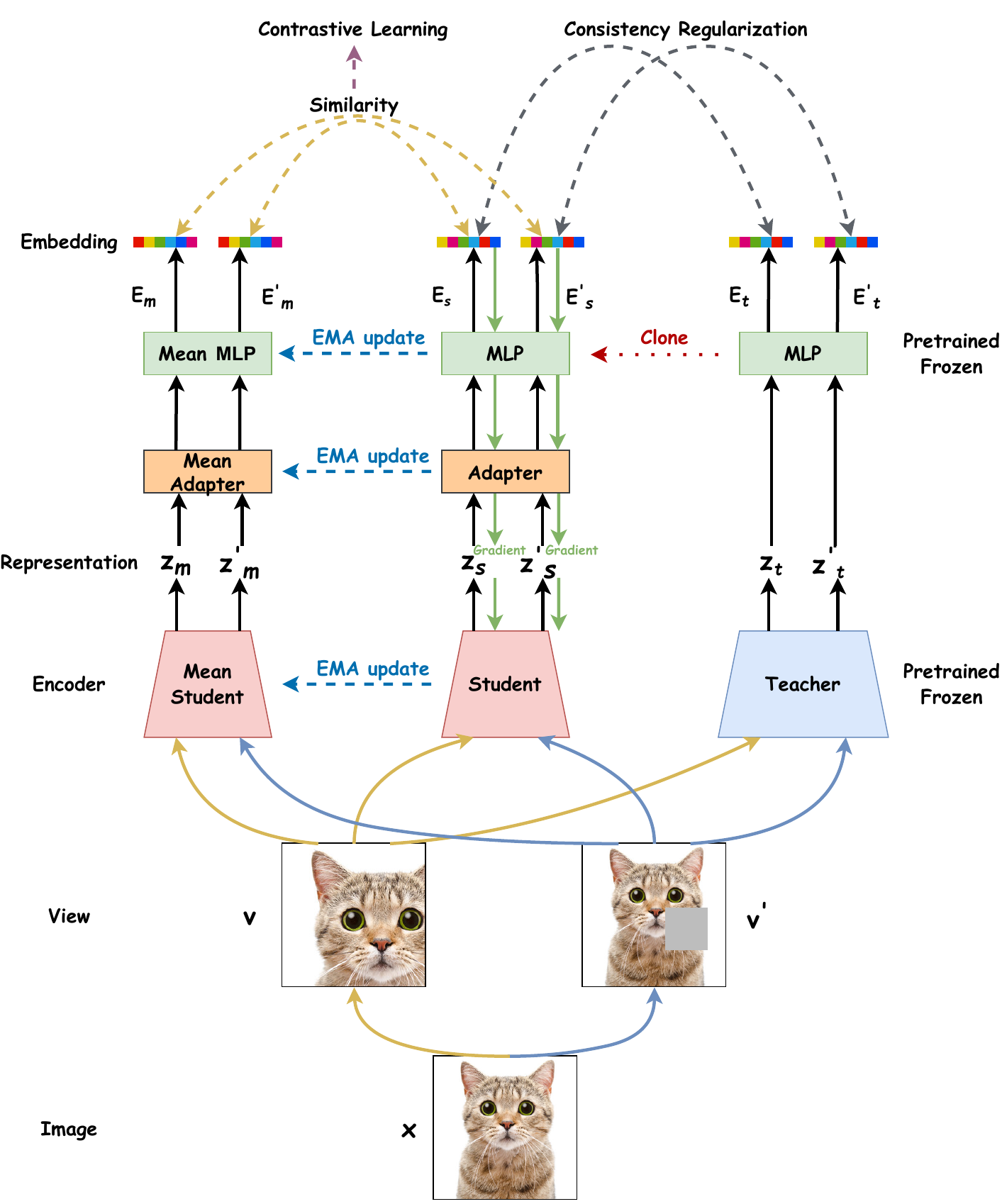}
    \caption{The pipeline of the proposed \textsc{Retro} technique.
    Two different data augmentation techniques first transform a single image into two views.
    A self-supervised pre-trained teacher is added in addition to the original contrastive SSL component, and the final embeddings generated by the learnable student and the frozen teacher must be consistent for each view.
    }
    \label{fig:pipeline}
\end{wrapfigure}

\subsubsection{DisCo}
In DisCo, the input $x$ is transformed into two views $v$ and $v'$ using two different augmentation strategies in each iteration. 
The views are fed into both the student encoder $f_s$ and the teacher encoder $f_t$ to create four representations $z_s$, $z'_s$, $z_t$, and $z'_t$.
These representations are then projected using two different projection heads $g_s$ and $g_t$ to produce low-dimensional embeddings $E_s$, $E'_s$, $E_t$, and $E'_t$, respectively. 
The same process is also applied with the mean student, resulting in representations $z_m$, $z'_m$ and embeddings $E_m$, $E'_m$. 
The embeddings are then used to compute the contrastive learning loss using InfoNCE loss \cite{oord2018representation}, similar to MoCo \cite{he2020momentum}, as follows:
\begin{equation}
\small
    \mathcal{L}_\text{con} = - \log \frac{\exp\left(\mathbf{q} \cdot \mathbf{k}^+ / \tau\right)}{\sum_{i=0}^{K} \exp\left(\mathbf{q} \cdot \mathbf{k}_i / \tau\right)},
\end{equation}
where $\mathbf{q}$ is the embedding $E_s$ of the student on view $v$, $\mathbf{k}$ is the embedding $E'_m$ of the mean student on view $v'$, $\tau$ is the temperature, and $K$ is the size of the memory bank. 
Additionally, the embeddings are used to compute a consistency loss using cosine similarity, which is represented using mean squared error (MSE) as follows:
\begin{equation}
\small
    \begin{gathered}
        \mathcal{L}_\text{dis} = \| E_s - E_t \|^2_2 + \| E'_s - E'_t \|^2_2
    \end{gathered}
\end{equation}

\subsection{\textsc{Retro}}
The overall framework of \textsc{Retro} is illustrated in Figure \ref{fig:pipeline}.
\textsc{Retro} comprises a lightweight student $s\left(\cdot\right)$, a mean student $m\left(\cdot\right)$, and a pre-trained frozen teacher $t\left(\cdot\right)$, which is similar to the DisCo framework \cite{gao2022disco}.
However, unlike DisCo, we propose that the pre-trained teacher projection head can be used directly for the students since it contains the most valuable knowledge.
Therefore, the objective is to train the student encoder to align the representation with the teacher projection head, instead of learning to mimic the teacher's behavior.
In addition, we leverage the power of the multi-view strategy by inputting both views into the mean student and enforcing the similarity constraint on pairs of embeddings between the student and the mean student.

To achieve this, we replace the student projection head with the teacher's projection head $g$, ensuring the consistency of all projection heads. 
However, since the input dimension of the student projection head is smaller than that of the teacher, we place an adapter $a\left(\cdot\right)$ between the encoder and projection head to align the dimension.
This process can be formulated as $E_s = g \circ a\left(z_s\right)$, $E'_s = g \circ a\left(z'_s\right)$, $E_t = g\left(z_t\right)$, $E'_t = g\left(z'_t\right)$, $E_m = g' \circ a'\left(z_m\right)$, and $E'_m = g' \circ a'\left(z'_m\right)$, where $a'\left(\cdot\right)$ and $g'\left(\cdot\right)$ are the mean adapter and mean projection head, respectively. These embeddings are then used to compute the contrastive loss and consistency loss.

\subsection{Loss function and parameter update process}
We follow BYOL \cite{grill2020bootstrap} to symmetrize the contrastive loss in \textsc{Retro} as follows:
\begin{equation}
\small
\begin{aligned}
    \mathcal{L}_\text{con} = \frac{1}{2}\left(- \log \frac{\exp\left(\mathbf{q} \cdot \mathbf{k^{'}}_{+} / \tau\right)}{\sum_{i=0}^{K} \exp\left(\mathbf{q} \cdot \mathbf{k^{'}}_{i} / \tau\right)}\right)
    +\frac{1}{2}\left(- \log \frac{\exp\left(\mathbf{q^{'}} \cdot \mathbf{k}_{+} / \tau\right)}{\sum_{i=0}^K \exp\left(\mathbf{q^{'}} \cdot \mathbf{k}_{i} / \tau\right)}\right)
\end{aligned}
\end{equation}
Here, $q$ and $q'$ are the embeddings from the student, while $k$ and $k'$ are the embeddings from the mean student. We use two different memory banks for the two different views $v$ and $v'$, respectively. 
The overall loss function of \textsc{Retro} is formulated as follows:
\begin{equation}
\small
    \mathcal{L} = \mathcal{L}_\text{dis} + \gamma \mathcal{L}_\text{con}
    \label{eq:loss}
\end{equation}
where $\mathcal{L}_\text{dis}$ is the consistency loss, and $\mathcal{L}_\text{con}$ is the contrastive loss of the conventional SSL method. $\gamma$ is a hyperparameter used to control the weight of the contrastive loss, which is typically set to $1$. The parameters of the student encoder are optimized using the objective from Equation \ref{eq:loss}, while the parameters of the entire mean student are updated using the exponential moving average strategy as follows:
\begin{equation}
    \theta_k \leftarrow m \theta_k + \left(1 - m\right) \theta_q
\end{equation}
Here, $m\in[0,1)$ is the momentum coefficient, which is typically set to $0.999$, and $\theta$ represents the model parameters.

\section{Experiments}
\subsection{Implementation Details}
We first pre-train the self-supervised teacher models on the ImageNet dataset \cite{russakovsky2015imagenet}, which contains $1.3$ million training images and $50,000$ validation images with $1,000$ categories, using the MoCo-V2 \cite{he2020momentum} framework.
We use ResNet as the backbone for the teacher models with different widths/depths, such as ResNet-50 (22.4M), ResNet-101 (40.5M), ResNet-152 (55.4M), and ResNet-50$\times$2 (94M).
We pre-train ResNet-50/101 using the MoCo-V2 framework for $200$ epochs, ResNet-152 for $800$ epochs, while ResNet-50$\times$2 is pre-trained using the SwAV \cite{caron2020unsupervised} method for $400$ epochs.

We evaluate our \textsc{Retro} method on five lightweight networks as students, including EfficientNet-B0 (4.0M), EfficientNet-B1 (6.4M), MobileNet-v3-Large (4.2M), ResNet-18 (10.7M), and ResNet-34 (20.4M).
We use the same learning rate and optimizer as MoCo-V2 and train all student models for $200$ epochs.
During distillation, the teacher's and student's projection heads are frozen for \textsc{Retro}, while SEED, DisCo, and BINGO only freeze the teacher.
As a result, \textsc{Retro} has fewer trainable parameters and simpler training objectives.
Note that SEED trains the models using a SwAV pre-trained teacher for $400$ epochs, and BINGO uses CutMix regularization \cite{yun2019cutmix} and more positive samples ($5\times$) during training, resulting in a higher benchmark score.
Additionally, BINGO is not an end-to-end framework.

We later fine-tune the self-supervised distillation models for linear evaluation on ImageNet for $100$ epochs.
We set the initial learning rate to $3$ for EfficientNet-B0/B1 and MobileNet-v3-Large, and $30$ for ResNet-18/34.
The learning rate is scheduled to decrease by a factor of $10$ at $60$ and $80$ epochs, and we use SGD as the optimizer.
We follow the other hyperparameters strictly as in MoCo-V2 \cite{he2020momentum}.

\subsection{Linear Evaluation}
The results presented in Table \ref{table:results-linear} demonstrate that students distilled by \textsc{Retro} outperform their counterparts pre-trained by MoCo-V2 (Baseline), and also outperform the prior state-of-the-art DisCo by a significant margin. 
However, we have not included CompRess in our comparison since it uses a teacher that was trained for $600$ epochs longer and distills for $400$ epochs longer than SEED, DisCo, and \textsc{Retro}. 
Therefore, it would be unfair to compare \textsc{Retro} to CompRess in this context.

\begin{table*}[!htb]
\centering
\resizebox{0.7\linewidth}{!}{
\begin{tabular}{c|c|cc|cc|cc|cc|cc}
\toprule    \toprule
\multirow{2}{*}{\textbf{Method}}  &   \multirow{2}{*}{\backslashbox{\textbf{T}}{\textbf{S}}} &   \multicolumn{2}{c|}{\textbf{Eff-b0}}  &   \multicolumn{2}{c|}{\textbf{Eff-b1}}  &   \multicolumn{2}{c|}{\textbf{Mob-v3}}  &   \multicolumn{2}{c|}{\textbf{R-18}}    &   \multicolumn{2}{c}{\textbf{R-34}}\\
\cmidrule(lr){3-12}
    &   &   \textbf{T-1} &   \textbf{T-5} &   \textbf{T-1} &   \textbf{T-5} &   \textbf{T-1} &   \textbf{T-5} &   \textbf{T-1} &   \textbf{T-5} &   \textbf{T-1} &   \textbf{T-5}\\   \midrule
\multicolumn{2}{c|}{\textbf{Supervised}}    &   77.1    &   93.3    &   79.2    &   94.4    &   75.2    &   -   &   72.1    &   -   &   75.0    &   -   \\  \midrule
\multicolumn{2}{c|}{\textbf{\textit{Self-supervised}}} &   &   &   &   &   &   &   &      \\
\multicolumn{2}{c|}{MoCo-V2 (Baseline)$\lozenge$} &   46.8    &   72.2    &   48.4    &   73.8    &   36.2    &   62.1    &   52.2    &   77.6    &   56.8    &   81.1    \\ \midrule
\textbf{\textit{SSL Distillation}}  &   &   &   &   &   &   &   &   &   &   &   \\
SEED \cite{fang2021seed}    &   R-50 (67.4)$\lozenge$ &   61.3    &   82.7    &   61.4    &   83.1    &   55.2    &   80.3    &   57.6    &   81.8    &   58.5    &   82.6    \\
DisCo \cite{gao2022disco}  &   R-50 (67.4)$\lozenge$   &   66.5 &   87.6 &   66.6 &   87.5 &   64.4 &   86.2 &   60.6  &   83.7  &   62.5  &   85.4  \\
BINGO \cite{xu2021bag}  &   R-50 (67.4)$\lozenge$   &   -   &   -   &   -   &   -   &   -   &   -   &   61.4  &   84.3  &   63.5  &   85.7  \\
\rowcolor{LightGreen} \textbf{\textsc{Retro}}    &   R-50 (67.4)$\lozenge$   &    \textbf{66.9}   &   \textbf{88.2}   &   \textbf{67.1}   &   \textbf{88.4}   &   \textbf{66.2}   &   \textbf{87.2}   &  \textbf{62.9}   &    \textbf{85.4}   & \textbf{64.1}   &   \textbf{86.8}   \\
    &   &   \textcolor{LightGreen}{($20.1\uparrow$)}    &   \textcolor{LightGreen}{($16.0\uparrow$)} &   \textcolor{LightGreen}{($18.7\uparrow$)} &   \textcolor{LightGreen}{($14.6\uparrow$)} &   \textcolor{LightGreen}{($30.0\uparrow$)} &   \textcolor{LightGreen}{($25.1\uparrow$)} &   \textcolor{LightGreen}{($10.7\uparrow$)}  &   \textcolor{LightGreen}{($7.8\uparrow$)}  &   \textcolor{LightGreen}{($7.3\uparrow$)}  &   \textcolor{LightGreen}{($5.7\uparrow$)}  \\ \midrule
SEED \cite{fang2021seed}    &   R-101 (70.3)    &   63.0    &   83.8    &   63.4    &   84.6    &   59.9    &   83.5    &   58.9    &   82.5    &   61.6    &   84.9    \\
DisCo \cite{gao2022disco}   &   R-101 (69.1)$\lozenge$   &   68.9 &   88.9 &   69.0 &   89.1 &   65.7 &   86.7 &   62.3 &   85.1  &   64.4  &   86.5  \\ 
\rowcolor{LightGreen} \textbf{\textsc{Retro}}    &   R-101 (70.3)   &    \textbf{69.3}   &   \textbf{89.8}   &   \textbf{69.9}   &   \textbf{89.9}   &   \textbf{67.5}   &   \textbf{88.6}   &  \textbf{64.8}   &    \textbf{86.6}   & \textbf{66.1}   &   \textbf{87.9}   \\
    &   &   \textcolor{LightGreen}{($22.5\uparrow$)}    &   \textcolor{LightGreen}{($17.6\uparrow$)} &   \textcolor{LightGreen}{($21.5\uparrow$)} &   \textcolor{LightGreen}{($16.1\uparrow$)} &   \textcolor{LightGreen}{($31.3\uparrow$)} &   \textcolor{LightGreen}{($26.5\uparrow$)} &   \textcolor{LightGreen}{($12.6\uparrow$)}  &   \textcolor{LightGreen}{($9.0\uparrow$)}  &   \textcolor{LightGreen}{($9.3\uparrow$)}  &   \textcolor{LightGreen}{($6.8\uparrow$)}  \\ \midrule
\midrule
SEED \cite{fang2021seed}    &   R-152 (74.2)    &   65.3    &   86.0    &   67.3    &   86.9    &   61.4    &   84.6    &   59.5    &   83.3    &   62.7    &   85.8    \\
DisCo \cite{gao2022disco}   &   R-152 (74.1)$\lozenge$   &   67.8 &   87.0 &   73.1 &   91.2 &   63.7 &   84.9 &   65.5 &   86.7  &   68.1 &   88.6  \\ 
BINGO \cite{xu2021bag}  &   R-152 (74.1)$\lozenge$   &   -  &   -   &   -   &   -   &   -   &   -   &   65.9  &   87.1  &   69.1  &   88.9  \\
\rowcolor{LightGreen} \textbf{\textsc{Retro}}    &   R-152 (74.1)$\lozenge$   &    \textbf{69.8}   &   \textbf{90.2}   &   \textbf{73.7}   &   \textbf{91.4}   &   \textbf{68.0}   &   \textbf{86.2}   &  \textbf{66.9}   &    \textbf{88.1}   & \textbf{69.4}   &   \textbf{89.9}   \\
    &   &   \textcolor{LightGreen}{($23.0\uparrow$)}    &   \textcolor{LightGreen}{($18.0\uparrow$)} &   \textcolor{LightGreen}{($25.3\uparrow$)} &   \textcolor{LightGreen}{($17.6\uparrow$)} &   \textcolor{LightGreen}{($31.8\uparrow$)} &   \textcolor{LightGreen}{($24.1\uparrow$)} &   \textcolor{LightGreen}{($14.7\uparrow$)}  &   \textcolor{LightGreen}{($10.5\uparrow$)}  &   \textcolor{LightGreen}{($12.6\uparrow$)}  &   \textcolor{LightGreen}{($8.8\uparrow$)}  \\ \midrule
\midrule
SEED \cite{fang2021seed}    &   R-50$\times$2 (77.3)$\dagger$   &   67.6    &   87.4    &   68.0    &   87.6    &   68.2    &   88.2    &   63.0    &   84.9    &   65.7    &   86.8    \\
DisCo \cite{gao2022disco}   &   R-50$\times$2 (77.3)$\dagger$   &   69.1 &   88.9 &   64.0 &   84.6 &   58.9 &   81.4 &   65.2   &   86.8  &   67.6 &   88.6  \\
BINGO \cite{xu2021bag}  &   R-50$\times$2 (77.3)$\dagger$   &   -  &   -   &   -   &   -   &   -   &   -   &   65.5  &   87.0  &   68.9  &   89.0  \\
\rowcolor{LightGreen} \textbf{\textsc{Retro}}    &   R-50$\times$2 (77.3)$\dagger$   &    \textbf{70.2}   &   \textbf{90.4}   &   \textbf{73.8}   &   \textbf{91.6}   &   \textbf{70.1}   &   \textbf{89.2}   &  \textbf{65.9}   &    \textbf{87.1}   & \textbf{68.9}   &   \textbf{89.7}   \\
    &   &   \textcolor{LightGreen}{($23.4\uparrow$)}    &   \textcolor{LightGreen}{($18.2\uparrow$)} &   \textcolor{LightGreen}{($25.4\uparrow$)} &   \textcolor{LightGreen}{($17.8\uparrow$)} &   \textcolor{LightGreen}{($33.9\uparrow$)} &   \textcolor{LightGreen}{($27.1\uparrow$)} &   \textcolor{LightGreen}{($13.7\uparrow$)}  &   \textcolor{LightGreen}{($9.5\uparrow$)}  &   \textcolor{LightGreen}{($12.1\uparrow$)}  &   \textcolor{LightGreen}{($8.6\uparrow$)}  \\
\bottomrule   \bottomrule
\end{tabular}%
}
\caption{ImageNet Test Accuracy (\%) using Linear Classification on Different Student Architectures.
In the table, $\lozenge$ indicates that the teacher and students are pre-trained with MoCo-V2, while $\dagger$ indicates that the teacher is pre-trained by SwAV. SEED distilled for $800$ epochs using R-50$\times$2 as the teacher, while DisCo, BINGO, and \textsc{Retro} distilled for $200$ epochs. "T" denotes the teacher, and "S" denotes the student. The subscript in green represents the improvement compared to the MoCo-V2 baseline.}
\label{table:results-linear}
\end{table*}

The results in Table \ref{table:results-linear} demonstrate that \textsc{Retro} outperforms prior methods SEED, DisCo, and BINGO across all benchmarked models. 
Notably, when using ResNet-50 as the teacher, \textsc{Retro} achieves state-of-the-art top-1 accuracy on all student models. 
Moreover, using ResNet-152 instead of ResNet-50 as the teacher leads to a significant improvement in the performance of ResNet-34, from $56.8\%$ to $69.4\%$. 
It is worth noting that when using \textsc{Retro} with ResNet-50/101 as the teacher, the linear evaluation result of EfficientNet-B0 is very close to that of the teacher, despite EfficientNet-B0 having only $9.4\%/16.3\%$ of the parameters of ResNet-50/101.

\begin{wraptable}{r}{0.5\linewidth}
    \centering
    \resizebox{0.8\linewidth}{!}{
    \begin{tabular}{lccc}
    \toprule \toprule
        \textbf{Method}         &   \textbf{T}  &   \textbf{1\% labels}&   \textbf{10\% labels}\\
        \midrule
        MoCo-V2 (Baseline)      &   -           &   30.9            &   45.8\\
        \midrule
        SEED\cite{fang2021seed} &   R-50 (67.4) &   39.1            &   50.2\\
        DisCo\cite{gao2022disco}&   R-50 (67.4) &   39.2            &   50.1\\
        BINGO\cite{xu2021bag}   &   R-50 (67.4) &   42.8            &   57.5\\
        \rowcolor{LightGreen} \textbf{\textsc{Retro}}    &   R-50 (67.4) &   \textbf{43.1}            &   \textbf{57.9}\\
        \midrule
        SEED\cite{fang2021seed} &   R-101 (70.3)&   41.4            &   54.8\\
        DisCo\cite{gao2022disco}&   R-101 (69.1)&   47.8            &   54.7\\
        \rowcolor{LightGreen} \textbf{\textsc{Retro}}    &   R-101 (70.3) &   \textbf{49.2}            &   \textbf{60.5}\\
        \midrule
        SEED\cite{fang2021seed} &   R-152 (74.1)&   44.3            &   54.8\\
        DisCo\cite{gao2022disco}&   R-152 (74.1)&   47.1            &   54.7\\
        BINGO\cite{xu2021bag}   &   R-152 (74.1)&   50.3            &   61.2\\
        \rowcolor{LightGreen} \textbf{\textsc{Retro}}    &   R-152 (74.1) &   \textbf{50.9}            &   \textbf{62.0}\\
        \midrule
        BINGO\cite{xu2021bag}   &   R-50$\times$2 (77.3)&  48.2            &   60.2\\
        \rowcolor{LightGreen} \textbf{\textsc{Retro}}    &   R-50$\times$2 (77.3) &   \textbf{50.6}            &   \textbf{61.9}\\
    \bottomrule \bottomrule
    \end{tabular}
    }
    \caption{Semi-supervised learning by fine-tuning 1\% and 10\% images on ImageNet using ResNet-18.}
    \label{tab:semi}
\end{wraptable}

\subsection{Semi-supervised Linear Evaluation}
We also evaluate our method in semi-supervised scenarios, following previous methodologies.
We use $1\%$ and $10\%$ sampled subsets of the ImageNet training data (i.e., $13$ and $128$ samples per class, respectively) to fine-tune the student models. As shown in Table \ref{tab:semi}, our \textsc{Retro} approach consistently outperforms the baseline under any quantity of labeled data. 
Notably, our method achieves these results while strictly following the settings from SEED \cite{fang2021seed} and DisCo \cite{gao2022disco}, whereas BINGO uses a higher learning rate ($10$) for the classifier layer.

Moreover, our experiments demonstrate that \textsc{Retro} is stable under varying percentages of annotations, indicating that students always benefit from being distilled by larger teacher models. 
The results also suggest that having more labeled data can help improve the final performance of the student models.

\begin{table*}[!htb]
\centering
\resizebox{0.6\linewidth}{!}{
\begin{tabular}{c|c|c|ccc|ccc|ccc}
\toprule    \toprule
\multirow{4}{*}{\textbf{S}}  &   \multirow{4}{*}{\textbf{T}} &   \multirow{4}{*}{\textbf{Method}}  &   \multicolumn{6}{c|}{\textbf{Object Detection}}  &   \multicolumn{3}{c}{\textbf{Instance Segmentation}}\\
\cmidrule(lr){4-12}
    &   &   &   \multicolumn{3}{c|}{\textbf{VOC}} &   \multicolumn{3}{c|}{\textbf{COCO}} &   \multicolumn{3}{c}{\textbf{COCO}}\\
\cmidrule(lr){4-12}
    &   &   &   \textbf{\textit{$AP^{bb}$}}   &    \textbf{\textit{$AP^{bb}_{50}$}}   & \textbf{\textit{$AP^{bb}_{75}$}}   &  \textbf{\textit{$AP^{bb}$}}   &   \textbf{\textit{$AP^{bb}_{50}$}}   &  \textbf{\textit{$AP^{bb}_{75}$}}   &  \textbf{\textit{$AP^{mk}$}}   &   \textbf{\textit{$AP^{mk}_{50}$}}   &  \textbf{\textit{$AP^{mk}_{75}$}}   \\   \midrule
\multirow{14}{*}{R-34}  &   $\times$    &   MoCo-V2   &   53.6   &    79.1   &    58.7   &    38.1   &    56.8   &    40.7   &    33.0   &    53.2   &    35.3   \\
\cmidrule(lr){2-12}
    &   \multirow{4}{*}{R-50}   &   SEED \cite{fang2021seed}    &   53.7 &   79.4    &   59.2    &   38.4    &   57.0    &   41.0    &   33.3    &   53.2    &   35.3    \\
    &   &   DisCo \cite{gao2022disco}  &   56.5   &   80.6 &   62.5 &   40.0 &   59.1 &   43.4 &   34.9 &   56.3  &   37.1  \\
    &   &   \cellcolor{LightGreen}\textbf{\textsc{Retro}}    &    \cellcolor{LightGreen}\textbf{57.2}   &   \cellcolor{LightGreen}\cellcolor{LightGreen}\textbf{81.4}   &   \cellcolor{LightGreen}\cellcolor{LightGreen}\textbf{63.3}   &   \cellcolor{LightGreen}\cellcolor{LightGreen}\textbf{41.5}   &   \cellcolor{LightGreen}\cellcolor{LightGreen}\textbf{60.2}   &   \cellcolor{LightGreen}\cellcolor{LightGreen}\textbf{45.3}   &   \cellcolor{LightGreen}\cellcolor{LightGreen}\textbf{35.9}   &   \cellcolor{LightGreen}\cellcolor{LightGreen}\textbf{57.8}   &   \cellcolor{LightGreen}\cellcolor{LightGreen}\textbf{38.8}   \\
    &   &   &   \textcolor{LightGreen}{($3.6\uparrow$)}    &   \textcolor{LightGreen}{($2.3\uparrow$)} &   \textcolor{LightGreen}{($4.6\uparrow$)} &   \textcolor{LightGreen}{($3.4\uparrow$)} &   \textcolor{LightGreen}{($3.4\uparrow$)} &   \textcolor{LightGreen}{($4.6\uparrow$)} &   \textcolor{LightGreen}{($2.9\uparrow$)}  &   \textcolor{LightGreen}{($4.6\uparrow$)} &   \textcolor{LightGreen}{($3.5\uparrow$)}    \\ 
\cmidrule(lr){2-12}
    &   \multirow{4}{*}{R-101}   &   SEED \cite{fang2021seed}    &   54.1 &   79.8    &   59.1    &   38.5    &   57.3    &   41.4    &   33.6    &   54.1    &   35.6    \\
    &   &   DisCo \cite{gao2022disco}  &   56.1   &   80.3 &   61.8 &   40.0 &   59.1 &   43.2 &   34.7 &   55.9  &   37.4  \\
    &   &   \cellcolor{LightGreen}\textbf{\textsc{Retro}}    &    \cellcolor{LightGreen}\cellcolor{LightGreen}\textbf{57.3}   &   \cellcolor{LightGreen}\textbf{81.8}   &   \cellcolor{LightGreen}\textbf{63.5}   &   \cellcolor{LightGreen}\textbf{41.5}   &   \cellcolor{LightGreen}\textbf{60.3}   &   \cellcolor{LightGreen}\textbf{45.4}   &   \cellcolor{LightGreen}\textbf{36.0}   &   \cellcolor{LightGreen}\textbf{57.8}   &   \cellcolor{LightGreen}\textbf{38.9}   \\
    &   &   &   \textcolor{LightGreen}{($3.7\uparrow$)}    &   \textcolor{LightGreen}{($2.7\uparrow$)} &   \textcolor{LightGreen}{($4.8\uparrow$)} &   \textcolor{LightGreen}{($3.4\uparrow$)} &   \textcolor{LightGreen}{($3.5\uparrow$)} &   \textcolor{LightGreen}{($4.7\uparrow$)} &   \textcolor{LightGreen}{($3.0\uparrow$)}  &   \textcolor{LightGreen}{($4.6\uparrow$)} &   \textcolor{LightGreen}{($3.6\uparrow$)}    \\ 
\cmidrule(lr){2-12}
    &   \multirow{4}{*}{R-152}   &   SEED \cite{fang2021seed}    &   54.4 &   80.1    &   59.9    &   38.4    &   57.0    &   41.0    &   33.3    &   53.7    &   35.3    \\
    &   &   DisCo \cite{gao2022disco}  &   56.6   &   80.8 &   63.4 &   39.4 &   58.7 &   42.7 &   34.4 &   55.4  &   36.7  \\
    &   &   BINGO \cite{xu2021bag}  &   -   &   -   &   -   &   39.9 &   59.4 &   43.5 &   35.7 &   56.5  &   38.2  \\
    &   &   \cellcolor{LightGreen}\textbf{\textsc{Retro}}    &    \cellcolor{LightGreen}\textbf{57.5}   &   \cellcolor{LightGreen}\textbf{81.9}   &   \cellcolor{LightGreen}\textbf{64.1}   &   \cellcolor{LightGreen}\textbf{41.4}   &   \cellcolor{LightGreen}\textbf{60.6}   &   \cellcolor{LightGreen}\textbf{45.4}   &   \cellcolor{LightGreen}\textbf{36.1}   &   \cellcolor{LightGreen}\textbf{57.3}   &   \cellcolor{LightGreen}\textbf{39.2}   \\
    &   &   &   \textcolor{LightGreen}{($3.9\uparrow$)}    &   \textcolor{LightGreen}{($2.8\uparrow$)} &   \textcolor{LightGreen}{($5.4\uparrow$)} &   \textcolor{LightGreen}{($3.3\uparrow$)} &   \textcolor{LightGreen}{($3.8\uparrow$)} &   \textcolor{LightGreen}{($4.7\uparrow$)} &   \textcolor{LightGreen}{($3.1\uparrow$)}  &   \textcolor{LightGreen}{($4.1\uparrow$)} &   \textcolor{LightGreen}{($3.8\uparrow$)}    \\ 
\bottomrule   \bottomrule
\end{tabular}%
}
\caption{Object detection and instance segmentation results on VOC-07 test and COCO val2017 using ResNet-34 as the backbone.
The subscript in green represents the improvement compared to the MoCo-V2 baseline.}
\label{table:results-detection-segmentation}
\end{table*}

\subsection{Transfer to CIFAR-10/CIFAR-100}
\begin{figure}[!ht]
    \centering
    \subfigure[ResNet-18]{\label{fig:cifar10-res18}\includegraphics[width=0.2\linewidth]{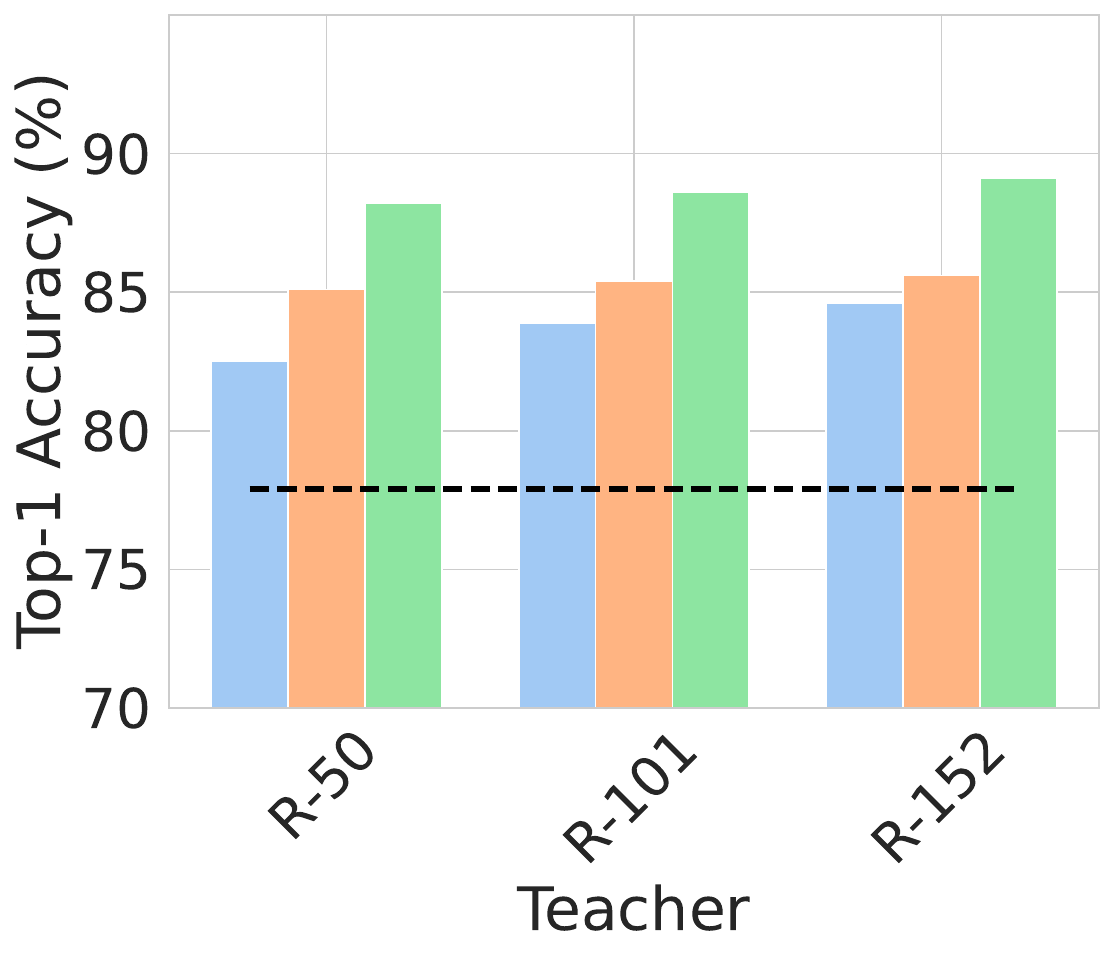}}
    \subfigure[EfficientNet-B0]{\label{fig:cifar10-effb0}\includegraphics[width=0.2\linewidth]{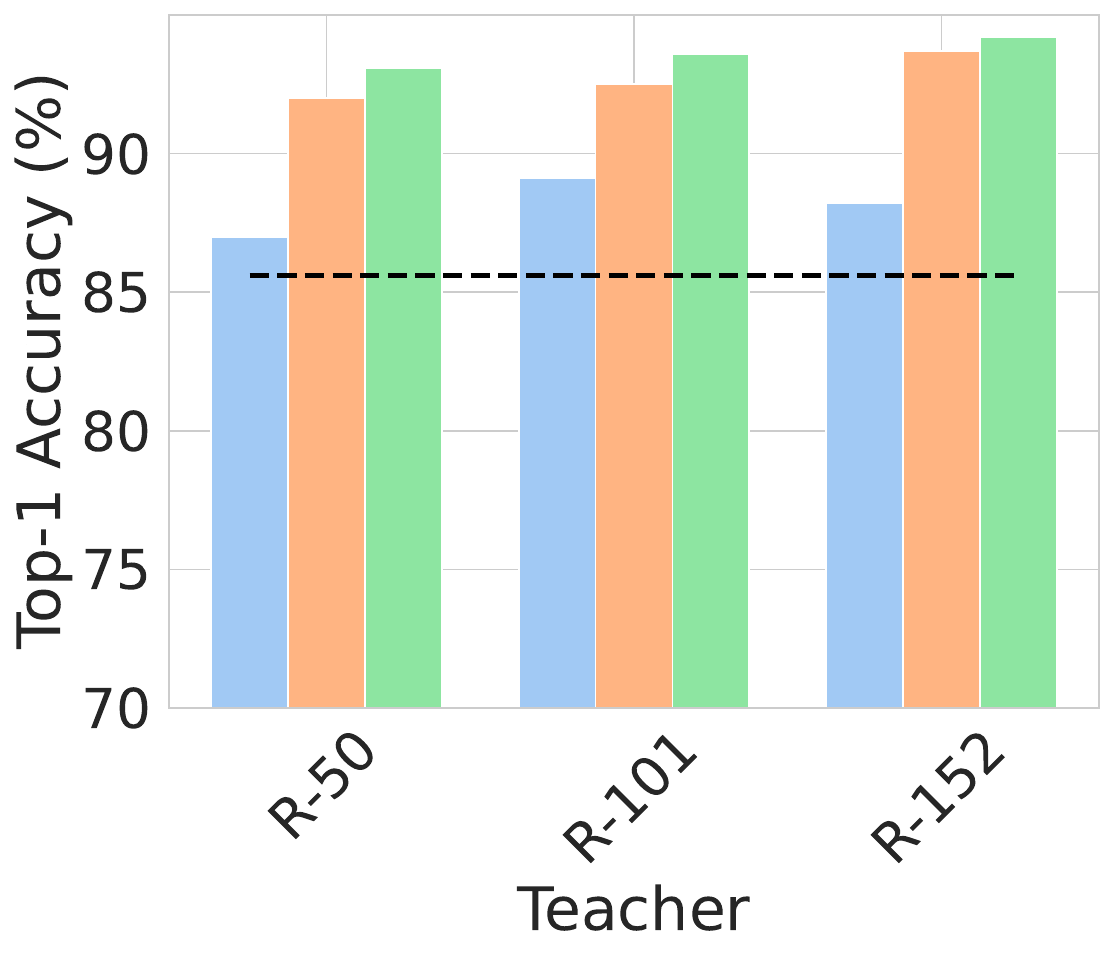}}
    \subfigure[ResNet-18]{\label{fig:cifar100-res18}\includegraphics[width=0.2\linewidth]{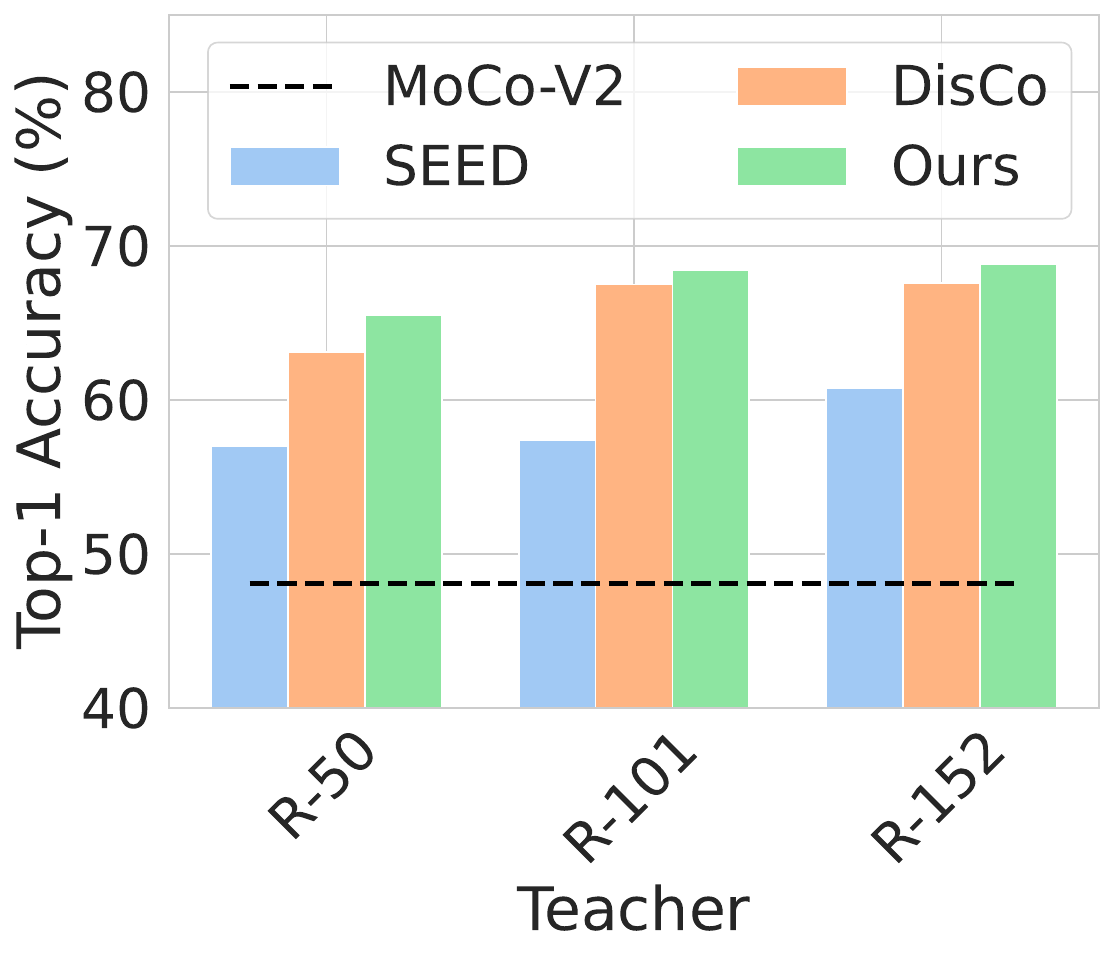}}
    \subfigure[EfficientNet-B0]{\label{fig:cifar100-effb0}\includegraphics[width=0.2\linewidth]{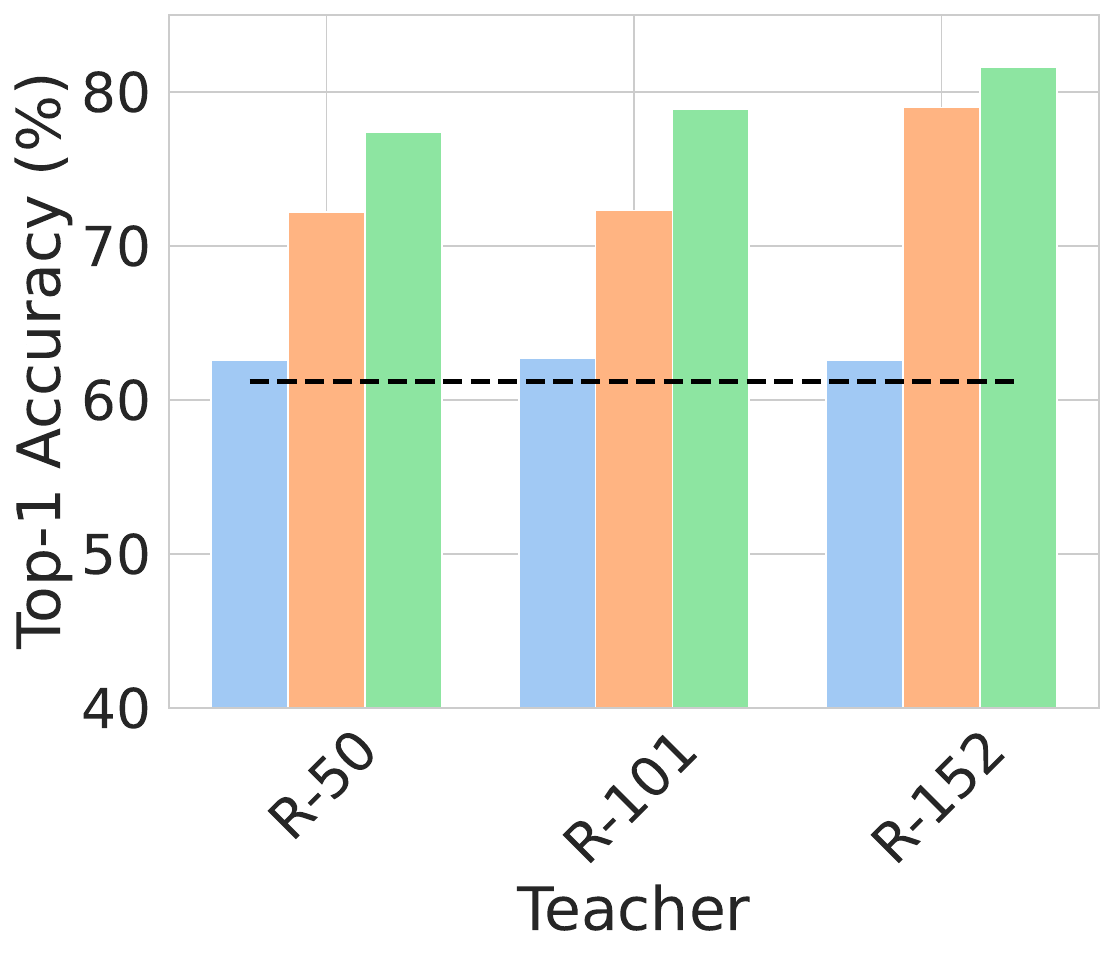}}
    
    \caption{Top-1 accuracy on CIFAR-10 (\ref{fig:cifar10-res18}, \ref{fig:cifar10-effb0}) and CIFAR-100 (\ref{fig:cifar100-res18}, \ref{fig:cifar100-effb0}) dataset.}
    \label{fig:cifar10}
\end{figure}

We conducted further evaluations to assess the generalization of representations obtained by \textsc{Retro} on CIFAR-10 and CIFAR-100 datasets, using ResNet-18/EfficientNet-B0 as a student and ResNet-50/ResNet-101/ResNet-152 as a teacher. 
The models were fine-tuned for $100$ epochs, with an initial learning rate of $3$ and the learning rate scheduler decreasing by a factor of $10$ at $60$ and $80$ epochs. 
All images were resized to $224\times224$, following the methodology from \cite{fang2021seed} since the original image resolution of the CIFAR dataset is $32\times32$. 
The results presented in Figure \ref{fig:cifar10} show that \textsc{Retro} outperforms prior methods SEED and DisCo across the datasets. 
Furthermore, the improvement brought by \textsc{Retro} becomes more apparent as the quality of the teacher models improves.

\subsection{Transfer to Detection and Segmentation}
For segmentation and downstream detection tasks, we adopt the fine-tuning methodology used in SEED \cite{fang2021seed} and DisCo \cite{gao2022disco}, where all parameters are fine-tuned. 
For the detection task on VOC \cite{everingham2015pascal}, the model is initialized with an initial learning rate of $0.1$, with $200$ warm-up iterations, and the learning rate is decreased by a factor of $10$ at $18$k and $22.2$k steps. 
The detector is trained for $48k$ steps, with a total batch size of $32$. 
During training, the input is randomly scaled from $400$ to $800$, and set to $800$ during inference. 
For detection and segmentation on COCO \cite{lin2014microsoft}, the model is trained for $180k$ iterations with an initial learning rate of $0.11$. 
During training, the image scale is randomized from $600$ to $800$.

We also performed tests on detection and segmentation tasks for generalization analysis.
Faster R-CNN \cite{ren2015faster} based on C4 is used for object detection for VOC and R-CNN Mask \cite{he2017mask} is used for object detection and version segmentation for COCO.
The results are displayed in Table \ref{table:results-detection-segmentation}.
In object detection, our method can yield clear improvements for both VOC and COCO datasets.
Also, as claimed by SEED \cite{fang2021seed}, the COCO training dataset has 118k images, while VOC has only 16.5k training images, so the improvement of COCO is relatively small compared to VOC. 
Therefore, the gain from initialization weight is relatively small.
\textsc{Retro} also has an advantage when it comes to instance segmentation tasks.

\section{Conclusion}
In this paper, we introduce the Reusing Teacher Projection head strategy (\textsc{Retro}), a novel approach for efficiently distilling self-supervised pre-trained teachers on lightweight models. 
Additionally, we impose symmetry contrastive learning to improve the representation between the student and mean student model. 
Despite its simplicity, our method outperforms prior methods by a large margin, demonstrating the importance of the projection head in distillation on lightweight models with fewer learnable parameters. 
\textsc{Retro} does not introduce any overhead during the inference phase. 
Our experiments show its superior performance across a range of architectures and tasks.

\bibliography{egbib}

\begin{thebibliography}{28}
\providecommand{\natexlab}[1]{#1}
\providecommand{\url}[1]{\texttt{#1}}
\expandafter\ifx\csname urlstyle\endcsname\relax
  \providecommand{\doi}[1]{doi: #1}\else
  \providecommand{\doi}{doi: \begingroup \urlstyle{rm}\Url}\fi

\bibitem[Caron et~al.(2020)Caron, Misra, Mairal, Goyal, Bojanowski, and
  Joulin]{caron2020unsupervised}
Mathilde Caron, Ishan Misra, Julien Mairal, Priya Goyal, Piotr Bojanowski, and
  Armand Joulin.
\newblock Unsupervised learning of visual features by contrasting cluster
  assignments.
\newblock \emph{Advances in neural information processing systems},
  33:\penalty0 9912--9924, 2020.

\bibitem[Caron et~al.(2021)Caron, Touvron, Misra, J{\'e}gou, Mairal,
  Bojanowski, and Joulin]{caron2021emerging}
Mathilde Caron, Hugo Touvron, Ishan Misra, Herv{\'e} J{\'e}gou, Julien Mairal,
  Piotr Bojanowski, and Armand Joulin.
\newblock Emerging properties in self-supervised vision transformers.
\newblock In \emph{Proceedings of the IEEE/CVF international conference on
  computer vision}, pages 9650--9660, 2021.

\bibitem[Chen et~al.(2020{\natexlab{a}})Chen, Kornblith, Norouzi, and
  Hinton]{chen2020simple}
Ting Chen, Simon Kornblith, Mohammad Norouzi, and Geoffrey Hinton.
\newblock A simple framework for contrastive learning of visual
  representations.
\newblock In \emph{International conference on machine learning}, pages
  1597--1607. PMLR, 2020{\natexlab{a}}.

\bibitem[Chen et~al.(2020{\natexlab{b}})Chen, Kornblith, Swersky, Norouzi, and
  Hinton]{chen2020big}
Ting Chen, Simon Kornblith, Kevin Swersky, Mohammad Norouzi, and Geoffrey~E
  Hinton.
\newblock Big self-supervised models are strong semi-supervised learners.
\newblock \emph{Advances in neural information processing systems},
  33:\penalty0 22243--22255, 2020{\natexlab{b}}.

\bibitem[Chen and He(2021)]{chen2021exploring}
Xinlei Chen and Kaiming He.
\newblock Exploring simple siamese representation learning.
\newblock In \emph{Proceedings of the IEEE/CVF conference on computer vision
  and pattern recognition}, pages 15750--15758, 2021.

\bibitem[Chen et~al.(2020{\natexlab{c}})Chen, Fan, Girshick, and
  He]{chen2020improved}
Xinlei Chen, Haoqi Fan, Ross Girshick, and Kaiming He.
\newblock Improved baselines with momentum contrastive learning.
\newblock \emph{arXiv preprint arXiv:2003.04297}, 2020{\natexlab{c}}.

\bibitem[Doersch et~al.(2015)Doersch, Gupta, and
  Efros]{doersch2015unsupervised}
Carl Doersch, Abhinav Gupta, and Alexei~A Efros.
\newblock Unsupervised visual representation learning by context prediction.
\newblock In \emph{Proceedings of the IEEE international conference on computer
  vision}, pages 1422--1430, 2015.

\bibitem[Everingham et~al.(2015)Everingham, Eslami, Van~Gool, Williams, Winn,
  and Zisserman]{everingham2015pascal}
Mark Everingham, SM~Ali Eslami, Luc Van~Gool, Christopher~KI Williams, John
  Winn, and Andrew Zisserman.
\newblock The pascal visual object classes challenge: A retrospective.
\newblock \emph{International journal of computer vision}, 111:\penalty0
  98--136, 2015.

\bibitem[Fang et~al.(2021)Fang, Wang, Wang, Zhang, Yang, and Liu]{fang2021seed}
Zhiyuan Fang, Jianfeng Wang, Lijuan Wang, Lei Zhang, Yezhou Yang, and Zicheng
  Liu.
\newblock Seed: Self-supervised distillation for visual representation.
\newblock \emph{arXiv preprint arXiv:2101.04731}, 2021.

\bibitem[Gao et~al.(2022)Gao, Zhuang, Lin, Cheng, Sun, Li, and
  Shen]{gao2022disco}
Yuting Gao, Jia-Xin Zhuang, Shaohui Lin, Hao Cheng, Xing Sun, Ke~Li, and
  Chunhua Shen.
\newblock Disco: Remedying self-supervised learning on lightweight models with
  distilled contrastive learning.
\newblock In \emph{Computer Vision--ECCV 2022: 17th European Conference, Tel
  Aviv, Israel, October 23--27, 2022, Proceedings, Part XXVI}, pages 237--253.
  Springer, 2022.

\bibitem[Gidaris et~al.(2018)Gidaris, Singh, and
  Komodakis]{gidaris2018unsupervised}
Spyros Gidaris, Praveer Singh, and Nikos Komodakis.
\newblock Unsupervised representation learning by predicting image rotations.
\newblock \emph{arXiv preprint arXiv:1803.07728}, 2018.

\bibitem[Grill et~al.(2020)Grill, Strub, Altch{\'e}, Tallec, Richemond,
  Buchatskaya, Doersch, Avila~Pires, Guo, Gheshlaghi~Azar,
  et~al.]{grill2020bootstrap}
Jean-Bastien Grill, Florian Strub, Florent Altch{\'e}, Corentin Tallec, Pierre
  Richemond, Elena Buchatskaya, Carl Doersch, Bernardo Avila~Pires, Zhaohan
  Guo, Mohammad Gheshlaghi~Azar, et~al.
\newblock Bootstrap your own latent-a new approach to self-supervised learning.
\newblock \emph{Advances in neural information processing systems},
  33:\penalty0 21271--21284, 2020.

\bibitem[He et~al.(2017)He, Gkioxari, Doll{\'a}r, and Girshick]{he2017mask}
Kaiming He, Georgia Gkioxari, Piotr Doll{\'a}r, and Ross Girshick.
\newblock Mask r-cnn.
\newblock In \emph{Proceedings of the IEEE international conference on computer
  vision}, pages 2961--2969, 2017.

\bibitem[He et~al.(2020)He, Fan, Wu, Xie, and Girshick]{he2020momentum}
Kaiming He, Haoqi Fan, Yuxin Wu, Saining Xie, and Ross Girshick.
\newblock Momentum contrast for unsupervised visual representation learning.
\newblock In \emph{Proceedings of the IEEE/CVF conference on computer vision
  and pattern recognition}, pages 9729--9738, 2020.

\bibitem[Hinton et~al.(2015)Hinton, Vinyals, and Dean]{hinton2015distilling}
Geoffrey Hinton, Oriol Vinyals, and Jeff Dean.
\newblock Distilling the knowledge in a neural network.
\newblock \emph{arXiv preprint arXiv:1503.02531}, 2015.

\bibitem[Koohpayegani et~al.(2020)Koohpayegani, Tejankar, and
  Pirsiavash]{abbasi2020compress}
Soroush~Abbasi Koohpayegani, Ajinkya Tejankar, and Hamed Pirsiavash.
\newblock Compress: Self-supervised learning by compressing representations.
\newblock \emph{Advances in neural information processing systems}, 2020.

\bibitem[Lin et~al.(2014)Lin, Maire, Belongie, Hays, Perona, Ramanan,
  Doll{\'a}r, and Zitnick]{lin2014microsoft}
Tsung-Yi Lin, Michael Maire, Serge Belongie, James Hays, Pietro Perona, Deva
  Ramanan, Piotr Doll{\'a}r, and C~Lawrence Zitnick.
\newblock Microsoft coco: Common objects in context.
\newblock In \emph{Computer Vision--ECCV 2014: 13th European Conference,
  Zurich, Switzerland, September 6-12, 2014, Proceedings, Part V 13}, pages
  740--755. Springer, 2014.

\bibitem[Noroozi and Favaro(2016)]{noroozi2016unsupervised}
Mehdi Noroozi and Paolo Favaro.
\newblock Unsupervised learning of visual representations by solving jigsaw
  puzzles.
\newblock In \emph{Computer Vision--ECCV 2016: 14th European Conference,
  Amsterdam, The Netherlands, October 11-14, 2016, Proceedings, Part VI}, pages
  69--84. Springer, 2016.

\bibitem[Oord et~al.(2018)Oord, Li, and Vinyals]{oord2018representation}
Aaron van~den Oord, Yazhe Li, and Oriol Vinyals.
\newblock Representation learning with contrastive predictive coding.
\newblock \emph{arXiv preprint arXiv:1807.03748}, 2018.

\bibitem[Park et~al.(2019)Park, Kim, Lu, and Cho]{park2019relational}
Wonpyo Park, Dongju Kim, Yan Lu, and Minsu Cho.
\newblock Relational knowledge distillation.
\newblock In \emph{Proceedings of the IEEE/CVF Conference on Computer Vision
  and Pattern Recognition}, pages 3967--3976, 2019.

\bibitem[Pathak et~al.(2016)Pathak, Krahenbuhl, Donahue, Darrell, and
  Efros]{pathak2016context}
Deepak Pathak, Philipp Krahenbuhl, Jeff Donahue, Trevor Darrell, and Alexei~A
  Efros.
\newblock Context encoders: Feature learning by inpainting.
\newblock In \emph{Proceedings of the IEEE conference on computer vision and
  pattern recognition}, pages 2536--2544, 2016.

\bibitem[Ren et~al.(2015)Ren, He, Girshick, and Sun]{ren2015faster}
Shaoqing Ren, Kaiming He, Ross Girshick, and Jian Sun.
\newblock Faster r-cnn: Towards real-time object detection with region proposal
  networks.
\newblock \emph{Advances in neural information processing systems}, 28, 2015.

\bibitem[Romero et~al.(2014)Romero, Ballas, Kahou, Chassang, Gatta, and
  Bengio]{romero2014fitnets}
Adriana Romero, Nicolas Ballas, Samira~Ebrahimi Kahou, Antoine Chassang, Carlo
  Gatta, and Yoshua Bengio.
\newblock Fitnets: Hints for thin deep nets.
\newblock \emph{arXiv preprint arXiv:1412.6550}, 2014.

\bibitem[Russakovsky et~al.(2015)Russakovsky, Deng, Su, Krause, Satheesh, Ma,
  Huang, Karpathy, Khosla, Bernstein, et~al.]{russakovsky2015imagenet}
Olga Russakovsky, Jia Deng, Hao Su, Jonathan Krause, Sanjeev Satheesh, Sean Ma,
  Zhiheng Huang, Andrej Karpathy, Aditya Khosla, Michael Bernstein, et~al.
\newblock Imagenet large scale visual recognition challenge.
\newblock \emph{International journal of computer vision}, 115:\penalty0
  211--252, 2015.

\bibitem[Tian et~al.(2019)Tian, Krishnan, and Isola]{tian2019contrastive}
Yonglong Tian, Dilip Krishnan, and Phillip Isola.
\newblock Contrastive representation distillation.
\newblock \emph{arXiv preprint arXiv:1910.10699}, 2019.

\bibitem[Xu et~al.(2020)Xu, Liu, Li, and Loy]{xu2020knowledge}
Guodong Xu, Ziwei Liu, Xiaoxiao Li, and Chen~Change Loy.
\newblock Knowledge distillation meets self-supervision.
\newblock In \emph{Computer Vision--ECCV 2020: 16th European Conference,
  Glasgow, UK, August 23--28, 2020, Proceedings, Part IX}, pages 588--604.
  Springer, 2020.

\bibitem[Xu et~al.(2021)Xu, Fang, Zhang, Xie, Wang, Dai, Xiong, and
  Tian]{xu2021bag}
Haohang Xu, Jiemin Fang, Xiaopeng Zhang, Lingxi Xie, Xinggang Wang, Wenrui Dai,
  Hongkai Xiong, and Qi~Tian.
\newblock Bag of instances aggregation boosts self-supervised distillation.
\newblock \emph{arXiv preprint arXiv:2107.01691}, 2021.

\bibitem[Yun et~al.(2019)Yun, Han, Oh, Chun, Choe, and Yoo]{yun2019cutmix}
Sangdoo Yun, Dongyoon Han, Seong~Joon Oh, Sanghyuk Chun, Junsuk Choe, and
  Youngjoon Yoo.
\newblock Cutmix: Regularization strategy to train strong classifiers with
  localizable features.
\newblock In \emph{Proceedings of the IEEE/CVF international conference on
  computer vision}, pages 6023--6032, 2019.

\end{thebibliography}

\clearpage
\appendix
\section{Impact of each contribution:}
In this section, we report the effectiveness of each of our contributions.
\ding{172} Reusing teacher projection head and \ding{173} symmetric contrastive learning loss.
We verify this via students that are trained with ResNet-50 as a teacher.

\begin{table}[!htb]
\centering
    \begin{small}
\resizebox{0.55\linewidth}{!}{
    \begin{tabular}{c|c|c|c|c|c}
    \toprule    \toprule
    \backslashbox{\textbf{Setting}}{\textbf{Student}\\\strut \textbf{Model}} &    \textbf{Eff-b0}  &   \textbf{Eff-b1}  &   \textbf{Mob-v3}  &   \textbf{R-18}    &   \textbf{R-34} \\
    \midrule    \midrule
    MoCo-V2 (Baseline)          & 46.8      & 48.4      & 36.2   & 52.2   & 56.8   \\ 
    \midrule
    DisCo \cite{gao2022disco}   & 66.5      & 66.6      & 64.4   & 60.6   & 62.5   \\ 
    + \ding{172}                & 66.7      & 66.9      & 65.8   & 62.5   & 63.4   \\ 
    \rowcolor{LightGreen}\textbf{+ \ding{173} (\textsc{Retro})}  & \textbf{66.9}      & \textbf{67.1}      & \textbf{66.2}  & \textbf{62.9}   & \textbf{64.1}   \\ 
    \bottomrule  \bottomrule
    \end{tabular}%
    }
    \caption{ImageNet top-1 accuracy (\%) using linear classification on different strategies.}
    \label{table:strategy}
    \end{small}
\end{table}

\section{Computational Complexity:}
As illustrated in Figure \ref{fig:pipeline}, the computational cost of \textsc{Retro} is higher compared to SEED \cite{fang2021seed} and DisCo \cite{gao2022disco} due to the additional forward propagation required for the mean student. 
The total number of forward propagation is $6$, which is three times higher than SEED and MoCo-V2. 
However, these additional forward propagations are not used during inference, so there is no overhead at inference time. 
The results in Table \ref{table:params} show that \textsc{Retro} has a lower number of learnable parameters than DisCo. Therefore, the run-time overhead of \textsc{Retro} is small and negligible compared to DisCo and BINGO. 
It should be noted that BINGO requires a KNN run to create a bag of positive samples, while \textsc{Retro} is an end-to-end approach.

\begin{table}[!htb]
    \centering
\resizebox{0.55\linewidth}{!}{
    \begin{tabular}{l|c|c|c|c|c}
        \toprule \toprule
        \textbf{Method} & \textbf{Eff-b0} & \textbf{Eff-b1} & \textbf{Mob-v3} & \textbf{R-18} & \textbf{R-34} \\
        \midrule \midrule
        DisCo \cite{gao2022disco} & 6.57M & 8.96M & 6.76M & 11.91M & 21.55M \\
        \rowcolor{LightGreen}\textbf{\textsc{Retro}} & 6.32M & 8.71M & 6.51M & 11.66M & 21.30M \\
        & \textcolor{LightGreen}{($\downarrow0.25$M)} & \textcolor{LightGreen}{($\downarrow0.25$M)} & \textcolor{LightGreen}{($\downarrow0.25$M)} & \textcolor{LightGreen}{($\downarrow0.25$M)} & \textcolor{LightGreen}{($\downarrow0.25$M)}\\
        \bottomrule \bottomrule
    \end{tabular}
}
    \caption{Comparison for the number of learnable parameters between DisCo and \textsc{Retro}.}
    \label{table:params}
\end{table}

\section{Comparison with other Distillation:}
\begin{table}[!htb]
\centering
\resizebox{0.35\linewidth}{!}{
    \begin{tabular}{lc}
    \toprule \toprule  
    \textbf{Method} & \textbf{Top-1} \\
    \midrule \midrule
    MoCo-V2 (Baseline) \cite{he2020momentum}        & 52.2 \\
    \midrule
    MoCo-V2 + KD \cite{fang2021seed}                & 55.3 \\
    MoCo-V2 + RKD \cite{park2019relational}         & 61.6 \\
    DisCo + KD \cite{gao2022disco}                  & 60.6 \\
    DisCo + RKD \cite{gao2022disco}                 & 60.6 \\
    BINGO \cite{xu2021bag}                          & 61.4 \\
    \rowcolor{LightGreen}\textbf{\textsc{Retro}}    & \textbf{62.9} \\
    \bottomrule \bottomrule
    \end{tabular}
}
\caption{Top-1 linear classification accuracy on ImageNet utilizing various distillation techniques on the ResNet-18 student model (ResNet-50 is used as teacher model).}
\label{table:distillations}
\end{table} 
For further verifying the strengths of \textsc{Retro}, we conducted the comparison against several different distillation strategies.
We include feature-based distillation (KD) and relation-based distillation (RKD), following DisCo \cite{gao2022disco} and BINGO \cite{xu2021bag}.
As shown in Table \ref{table:distillations}, \textsc{Retro} shows superior performance compared with other distillation methods and surpasses them by a large margin.

\section{Adapter settings}
\begin{figure}[!ht]
    \centering
    \subfigure[ResNet]{\includegraphics[width=0.3\linewidth]{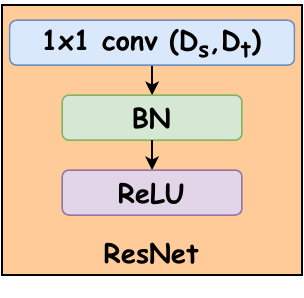}}
    \subfigure[EfficientNet]{\includegraphics[width=0.3\linewidth]{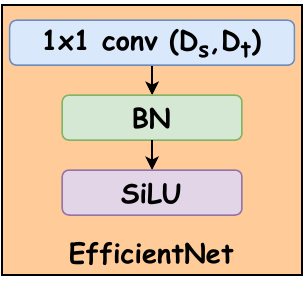}}
    \subfigure[MobileNet-v3]{\includegraphics[width=0.3\linewidth]{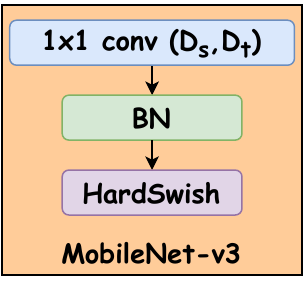}}
    
    \caption{Adapter structure for different student networks.}
    \label{fig:adapter}
\end{figure}

We visualize the adapter architecture for different student networks in Figure \ref{fig:adapter}.
The adapter for ResNet, EfficientNet, and MobileNet-v3 is a 1-D convolution layer that receives output from the student encoder ($D_s$) and aligns that for the teacher projection head ($D_t$), follows by a batch normalization and a non-linear activation layer.
Note that the adapter is placed right before the last pooling layer.

\section{Frozen vs unfrozen projection head}
We conduct the comparison between training \textsc{Retro} for $200$ epochs with frozen projection head and training for $170$ epochs with frozen and $30$ epochs with unfrozen projection head.
As we can see from Table \ref{table:training}, the latter training scheme produces better performance.

\begin{table}[!htb]
    \centering
\resizebox{0.55\linewidth}{!}{
    \begin{tabular}{l|c|c|c|c|c}
        \toprule \toprule
        \textbf{Method} & \textbf{Eff-b0} & \textbf{Eff-b1} & \textbf{Mob-v3} & \textbf{R-18} & \textbf{R-34} \\
        \midrule \midrule
        $170$/$30$ frozen/unfrozen epochs & 66.2 & 66.7 & 65.8 & 62.1 & 63.8 \\
        \rowcolor{LightGreen} $200$ frozen epochs & 66.9 & 67.1 & 66.2 & 62.9 & 64.1 \\
        \bottomrule \bottomrule
    \end{tabular}
}
    \caption{Comparison for the training scheme of \textsc{Retro}.}
    \label{table:training}
\end{table}

\end{document}